\documentclass{article} 
\usepackage[submission]{colm2025_conference}

\usepackage{microtype}
\usepackage{hyperref}
\usepackage{url}
\usepackage{booktabs}

\usepackage{lineno}

\definecolor{darkblue}{rgb}{0, 0, 0.5}
\hypersetup{colorlinks=true, citecolor=darkblue, linkcolor=darkblue, urlcolor=darkblue}

\usepackage{bbm}
\usepackage{amssymb} 
\usepackage{amsmath}
\usepackage{booktabs}   
\usepackage{caption}    
\usepackage[most]{tcolorbox}
\usepackage{multicol}
\usepackage{lipsum} 
\usepackage{indentfirst}
\usepackage[utf8]{inputenc}
\usepackage{multirow} 
\usepackage{algorithm}
\usepackage[noend]{algorithmic}  
\usepackage[algo2e]{algorithm2e} 
\usepackage{amsmath} 
\usepackage{wrapfig}
\usepackage{xcolor}
\usepackage{setspace}  
\usepackage[most]{tcolorbox}
\usepackage{booktabs}
\usepackage{adjustbox}
\definecolor{darkgreen}{RGB}{0,160,0}
\usepackage{graphicx} 
\usepackage[absolute,overlay]{textpos}
\usepackage{lipsum} 
\newcommand{\yes}{\checkmark}
\newcommand{\no}{$\times$}

\title{Exploring Expert Failures Improves LLM Agent Tuning}

\author{
\textbf{Li-Cheng Lan}\textsuperscript{1} \quad \quad \textbf{Andrew Bai}\textsuperscript{1} \quad \quad \textbf{Minhao Cheng}\textsuperscript{2} 
  \vspace{2pt}\\
   \textbf{Cho-Jui Hsieh}\textsuperscript{1} \quad \quad \textbf{Tianyi Zhou}\textsuperscript{4} \vspace{10pt}\\
  \textsuperscript{1}UCLA \quad \textsuperscript{2}Pennsylvania State University  \quad \textsuperscript{4}University of Maryland \vspace{5pt}\\
  \texttt{lclan@cs.ucla.edu \enskip andrewbai@ucla.edu \enskip mmc7149@psu.edu} \\ 
  \texttt{chohsieh@cs.ucla.edu \enskip tianyi@umd.edu} 
}

\colmfinaltrue
\colmsubmissionfalse
\colmpreprintfalse

\begin{document}

\ifcolmsubmission
\linenumbers
\fi

\maketitle

\begin{abstract}
Large Language Models (LLMs) have shown tremendous potential as agents, excelling at tasks that require multiple rounds of reasoning and interactions. 
Rejection Sampling Fine-Tuning (RFT) has emerged as an effective method for finetuning LLMs as agents: it first imitates expert-generated successful 
trajectories and further improves agentic skills through iterative fine-tuning on successful, self-generated trajectories.
However, since the expert (e.g., GPT-4) succeeds primarily on simpler subtasks and RFT inherently favors simpler scenarios, many complex subtasks remain unsolved and persistently out-of-distribution (OOD).
Upon investigating these challenging subtasks, we discovered that previously failed expert trajectories can often provide valuable guidance, e.g., plans and key actions, that can significantly improve agent exploration efficiency and acquisition of critical skills.
Motivated by these observations, we propose Exploring Expert Failures (EEF), which identifies beneficial actions from 
failed expert trajectories and integrates them into the training dataset. 
Potentially harmful actions are meticulously excluded to prevent contamination of the model learning process.
By leveraging the beneficial actions in expert failures, EEF successfully solves some previously unsolvable subtasks and improves agent tuning performance. 
Remarkably, our approach achieved a 62\% win rate in WebShop, outperforming RFT (53. 6\%) and GPT-4 (35. 6\%), and to the best of our knowledge, setting a new state-of-the-art as the first method to surpass a score of 0.81 in WebShop and exceed 81 in SciWorld.  \looseness-1

\end{abstract}

\section{Introduction}
\label{sec:intro}

Large Language Models (LLMs) have recently demonstrated remarkable potential as autonomous agents, extending beyond their traditional role as passive text generators~\citep{achiam2023gpt,team2023gemini,  zheng2023judging, xi2025rise}.
By allowing LLM-based agents to interact with their environment, they can autonomously complete real-world tasks by processing feedback, maintaining context, and making informed decisions~\citep{zhou2023webarena, yao2022webshop}.

To develop lightweight specialized agents, researchers commonly employ Rejection Sampling Fine-Tuning (RFT)~\citep{Touvron2023Llama2O, yuan2023scaling, xi2024agentgym}, which fine-tunes models using only environment-validated positive expert and synthesized trajectories.
Initially, an expert model (e.g., GPT-4~\citep{achiam2023gpt}) generates trajectories for each subtask. A smaller model then undergoes supervised fine-tuning (SFT) on only the positive expert trajectories. The fine-tuned model explores further, collecting additional successful trajectories to augment the training dataset. Repeated cycles of exploration and fine-tuning allow the smaller model to eventually exceed the expert's performance.
RFT is simple and scalable, avoiding the complexity of reward model training and hyperparameter tuning~\citep{Touvron2023Llama2O, yuan2023scaling}.
\begin{wrapfigure}{R}{0.45\textwidth}
\vspace{-1.em}
  \centering
  \includegraphics[width=0.45\textwidth]{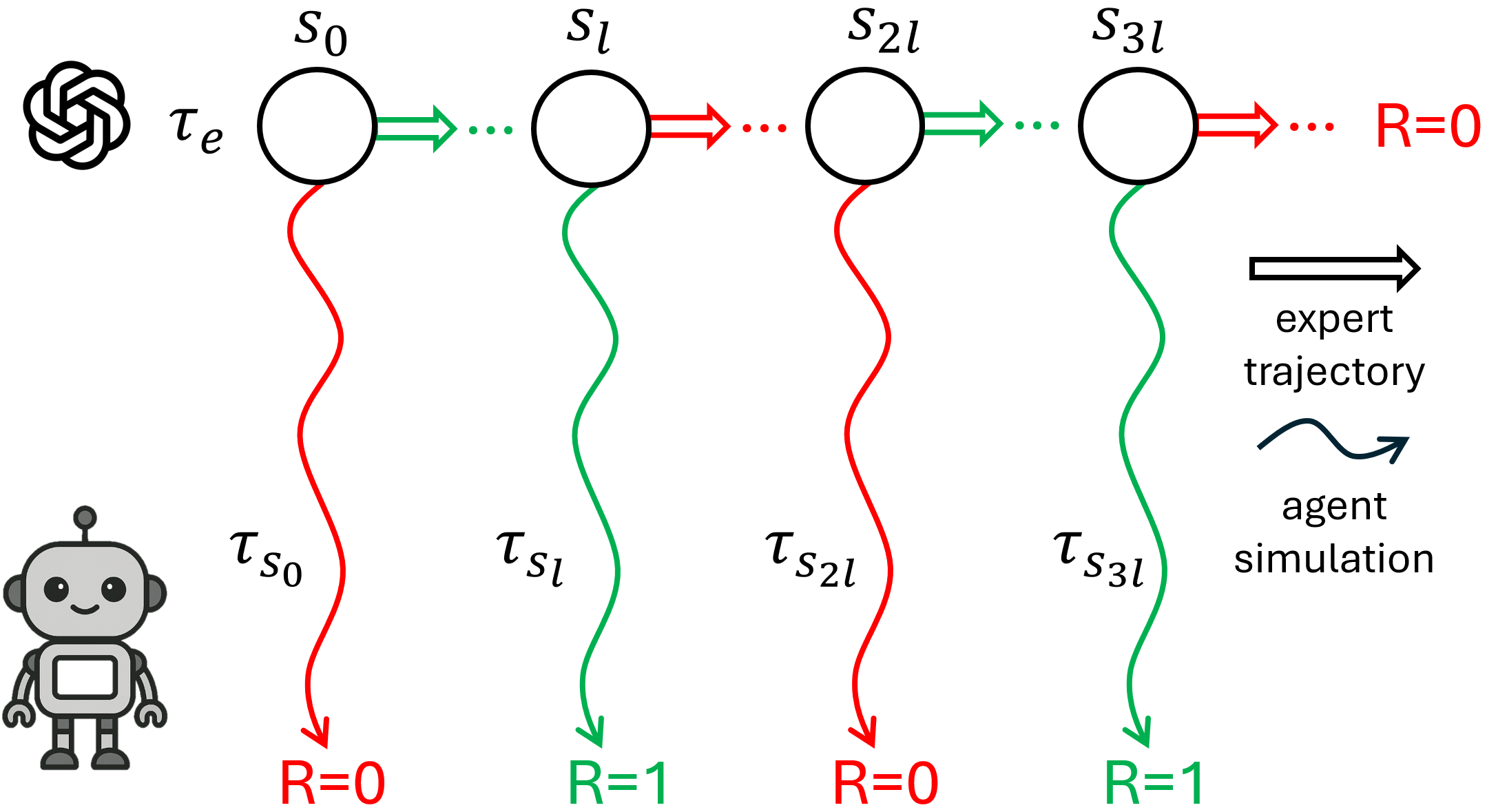}
  \caption{
\(\tau_e=[s_0, s_1, \dots]\) is a failed expert (GPT-4) trajectory for a challenging subtask \(s_0\) with \( R(\tau_e) = 0 \).  
Despite its overall failure, \(\tau_e\) contains partial insights to train agent $\pi_\theta$.
To identify beneficial actions in \(\tau_e\), our method (EEF) performs simulations from expert states \([s_0, s_l, \dots]\) at intervals of length \( l \), resulting in \([\tau_{s_0}, \tau_{s_l}, \dots]\), revealing that certain expert actions $a_{0:l-1}$ and $a_{2l:3l-1}$ 
enable the agent to transit from failed states \(s_0, s_{2l}\) to positively-performing states \(s_l, s_{3l}\). 
EEF thus considers these actions  ({\color{darkgreen}green} arrows) beneficial for solving  \(s_0, s_{2l}\) and retains them in the SFT training dataset while rejecting the remaining actions ({\color{red}red}  arrows).
}
\vspace{-.5em}
  \label{fig:motivation}
\end{wrapfigure}

However, in challenging environments, RFT can quickly be trapped in local minima, impeding further improvements in model performance. \cite{tong2024dart} demonstrates that this occurs primarily because RFT tends to favor simpler subtasks during data synthesis, leading to datasets biased toward easy scenarios with inadequate coverage of more difficult tasks. This issue is exacerbated in highly challenging environments, where even expert models struggle to solve the majority of subtasks. Consequently, these challenging subtasks remain Out-of-Distribution (OOD) throughout training. 
The issue of diminishing returns on harder subtasks has also been reported by other researchers~\citep{yuan2023scaling, Singh2023BeyondHD}, highlighting a key obstacle in training LLM agents.

To this end, we investigate the rejected, negative expert trajectories on these hard subtasks. We surprisingly found that due to the expert's stronger generalized capabilities,
the failed expert trajectories still consist of helpful actions or thoughts to solve the problem. The failure may stem from mistakes in the actions of the last few steps. 
This observation motivates us to take advantage of the numerous failed expert trajectories that may still contain beneficial actions.
Although prior works such as ETO~\citep{song2024trial} (based on DPO~\citep{Rafailov2023DirectPO}) and NAT~\citep{wang2024learning} are capable of utilizing negative data, they treat all actions within failed trajectories as uniformly negative, thereby failing to learn from potentially valuable actions embedded in them directly.

To learn these beneficial actions in negative trajectories, we proposed Exploring Expert Failures (``EEF''), which simulates the selected expert states and analyzes the simulation results to identify the beneficial actions and thoughts we observed for further fine-tuning. An illustration of our idea is demonstrated in Fig.~\ref{fig:motivation}. As shown in Fig.~\ref{fig:motivation}, state \(s_0\) is an initial state of a challenging subtask where our agent and the expert did not succeed. However, with the help of the negative expert trajectory \(\tau_e\), our agent can achieve success by starting simulations from experts' intermediate states \(s_l, s_{3l}\).
Based on the simulation results, we train the agent policy on those beneficial actions (green arrows).
Moreover, since our method exclusively learns from pinpointed beneficial actions, it enables us to utilize trajectories generated by weaker yet more cost-effective experts, such as GPT-3.5 Turbo, without being negatively influenced by their suboptimal actions.

We evaluated our method in Webshop and SciWorld, two highly challenging environments where even expert agents frequently fail. Our results demonstrate that effectively leveraging negative expert demonstrations enables our approach to solve subtasks previously unattainable by RFT. Consequently, our method outperforms baselines while retaining RFT’s simplicity. Notably, we are the first to surpass a 0.81 score in Webshop and to exceed a score of 81 in SciWorld.

\section{Backgroud}
\label{sec:background}

\textbf{Environment for text-based LLM agents.}
Similar to reinforcement learning (RL), the environment of text-based LLM agents can be considered as a Markov Decision Process (MDP) with contexts (subtasks)~\citep{Hallak2015ContextualMD}, formalized as the tuple \(M = \langle  \mathcal{S}, \mathcal{A}, T, R, \mathcal{C}\rangle\). 
The state space \(\mathcal{S}\) consists of states \(s \in \mathcal{S}\), each encoding both observable and unobservable aspects of the current status, as well as the dialogue history.
Note that the most recent message in the dialogue serves as the observation of \(s_0\).
The action space \(\mathcal{A}\) encompasses all possible actions the agent can perform, explicitly including internal reasoning steps (e.g., thoughts~\citep{Wei2022ChainOT}).
The transition function \(T: \mathcal{S} \times \mathcal{A} \times \mathcal{S} \to \mathbb{R}\) specifies the probability of transiting to state $s_{t+1}$ after taking action $a_t$ in state $s_t$. 
The reward function \(R: \mathcal{S} \times \mathcal{A} \to \mathbb{R}\) assigns a numerical reward to each state-action pair. In this paper, we consider a sparse reward structure, meaning the agent receives a nonzero reward only upon completing a subtask, while intermediate steps yield zero reward.
For convenience, give a trajectory $\tau = \bigl[s_0, a_0, s_1, a_1, \dots, s_T\bigr]$, we let \(R(\tau)\) denote the final total reward obtained by a trajectory \(\tau\).
We define the context space \( \mathcal{C} \subset \mathcal{S} \) as a set of initial states. Each initial state \( s_0 \in \mathcal{C} \) corresponds to a specific subtask defined by both an instruction and the current environmental situation. 
Note that even with the same instruction (query), the subtasks \( s_0 \in \mathcal{C} \) may be different due to unobservable parts of their states.
When developing an agent, we partition the subtasks into training set \( \mathcal{C}_{\text{train}}  \subset  \mathcal{C}\) and testing set \( \mathcal{C}_{\text{test}} \subset  \mathcal{C}\). During training, the agent can only explore the subtasks \(s_0\) in \( \mathcal{C}_{\text{train}}\). In addition, we assume that setting the environment state is permitted, allowing simulations to start from arbitrary states. This is a common practice in many RL environments for exploring OOD states~\citep{Lan2023CanAR}.

\textbf{Rejection Sampling Fine-Tuning (RFT).} 
RFT~\citep{yuan2023scaling} is a training paradigm that begins with expert demonstrations and then iteratively refines the learned policy through exploration.
At the beginning, the agent \(\pi_{\theta}\) is trained exclusively on the positive trajectories \(R(\tau_e) = 1\) 
of an expert dataset \(\mathcal{D}_e\). 
This initial phase helps the agent acquire fundamental skills and strategies. 
In subsequent iterations, for each subtask \(s_0\), the current policy generates \(k\) trajectories, and only the positive ones are added to the training set for further fine-tuning. 
Through repeated exploration-and-fine-tuning cycles, RFT not only exceeds expert-level performance but can also tackle subtasks beyond the capabilities of the original expert~\citep{song2024trial}. 
Owing to its simplicity and demonstrable effectiveness, we adopt RFT as our foundational framework, while focusing on refining the exploration process and improving the selection of successful trajectories for fine-tuning.



\begin{wrapfigure}{R}{0.4\textwidth}
\vspace{-15pt} 
\begin{tcolorbox}[title={\scriptsize \textbf{Box 1}: WebShop— shopping across over a million real-world Amazon products.}, colback=blue!5!white, colframe=blue!75!black, enhanced, boxed title style={colframe=blue!75!black,colback=blue!20!white}]
    \scriptsize
    \setlength{\parindent}{0pt} \textbf{Human instruction}:
    \raggedright
    \setlength{\parindent}{25pt} I need a m-size  white t-shirt under 40 dollars. \\[0.2\baselineskip]
    
    \setlength{\parindent}{0pt} \textbf{AI}:\ \ \ \  search[white t-shirt] \\[0.2\baselineskip]
    
    \setlength{\parindent}{0pt} \textbf{Obs}: [Next]

    \setlength{\parindent}{17pt} [A] Nike white t-shirt, 90 usd. 
    
    \setlength{\parindent}{17pt} [B] Addidas white t-shirt, 60 usd.
    
    \setlength{\parindent}{17pt} [C] Zara white t-shirt, 70 usd. \\[0.2\baselineskip]

    \setlength{\parindent}{0pt} \textbf{AI}:\ \ \ \    All too expensive, click[Next] \\[0.2\baselineskip]

    \setlength{\parindent}{0pt} \textbf{Obs}: [Next] [Prev]
    
    \setlength{\parindent}{17pt} [D] H\&M white t-shirt, 30 usd. 
    
    \setlength{\parindent}{17pt} [E] F21 black t-shirt, 40 usd.
    
    \setlength{\parindent}{17pt} [F] Uniqlo white jacket, 35 usd. \\[0.2\baselineskip]
    
   \setlength{\parindent}{0pt} \textbf{AI }:\ \ \ \   Only D is a white t-shirt, click[D] \\[0.2\baselineskip]

    \textbf{Obs}: [back to search]
    
    \setlength{\parindent}{17pt} size: [small] [median] [large]

    \setlength{\parindent}{0pt} \textbf{AI}:\ \ \ \   click[median] \\[0.2\baselineskip]

    \setlength{\parindent}{0pt} \textbf{Obs}: [buy now] [back to product] \\[0.2\baselineskip]
    
    \setlength{\parindent}{0pt} \textbf{AI}:\ \ \ \   click[buy now] \\[0.2\baselineskip]

    \textbf{Obs}: finished

\end{tcolorbox}
\vspace{-10pt} 
\end{wrapfigure}

\textbf{Webshop.}  
Webshop is an e-commerce simulation platform that replicates real-world online shopping behaviors by leveraging over a million Amazon product listings. 
We will later utilize Webshop to illustrate our motivation.
\textbf{Box 1} provided a successful trajectory of Webshop. Starting from a human instruction,
the agent then search (search[white t-shirt]), navigate (click[Next]), select (click[D], click[median]), and purchase ([buy now]) the correct item. 
This example shows that Webshop is a valuable environment for studying grounded language understanding and multistep decision-making, as it provides realistic, interactive subtasks that require interpreting language and executing sequential actions.

One of Webshop's key testing capabilities is assessing an agent's navigation skill, which is also crucial for many other web tasks~\citep{yao2022webshop}. 
Webshop provides two essential navigation actions: \textbf{Next} and \textbf{Back}. The \textbf{Next} action (i.e., Action:click[Next]) allows the agent to access subsequent pages with additional products (as illustrated in \textbf{Box 1}), while the \textbf{Back} action (i.e., Action:click[back to search]) enables the agent to discard the currently selected product and restart the search.
Learning these skills helps agent recover from a mistake or explore more candidates before making a decision, especially in challenging subtasks that even the expert fails to succeed at.

\section{Methodology}

\subsection{Motivation}
\label{sec:motivation}
In this subsection, we leverage observations from a preliminary experiment of Webshop to elucidate the key motivations behind our proposed approach.
In this preliminary experiment, we apply RFT to a small model, LLama3 8B~\citep{llama3modelcard}, with an expert dataset \(D_e\) generated by GPT-4.
It is important to note that the WebShop task is highly challenging, with GPT-4 only achieving approximately 35\% success on the training subtasks \(\mathcal C_{\text{train}}\). Consequently, the remaining \(65\%\) of subtasks lack expert demonstrations and remain out-of-distribution (OOD) after the first iteration of fine-tuning. Although subsequent iterations of RFT allow our agent to generalize and explore successful trajectories for some previously unsolved OOD subtasks, roughly 50\% of these subtasks remain unsolved and thus persist as OOD. 
We regard this as a waste, as these OOD subtasks and their expert trajectories are accessible yet remain unutilized by our agent during training. 

To solve the remaining \(50\%\) of unsolved subtasks, we analyze the failure trajectories produced by both our model and the expert.  In many of these cases, we observe that our agent fails to take necessary navigation actions such as \textbf{Next} and \textbf{Back}. For instance (\textbf{Box 1}), when the listed product does not meet the requirements (too expensive), our agent still blindly selects one of the options instead of using the \textbf{Next} action to check the next pages for more products.
Upon inspecting the expert trajectories—though they also result in failure—we find that the expert does attempt to use the \textbf{Next} action appropriately, but fails due to other factors. We even observed that our policy can succeed on some of the previously unsolved tasks when initialized from certain expert states, showing that those expert actions are truly beneficial to our agent.


\begin{wrapfigure}{R}{0.6\textwidth}
\vspace{-15pt} 
\begin{tcolorbox}[title={\scriptsize \textbf{Box 2}: 
A negative GPT-3 trajectory. 
GPT-3 first makes a mistake (red), followed by a recovery action (green).}, colback=blue!5!white, colframe=blue!75!black, enhanced, boxed title style={colframe=blue!75!black,colback=blue!20!white}]
    \scriptsize
    \setlength{\parindent}{0pt} \textbf{Human instruction}:
    \raggedright
    \setlength{\parindent}{25pt} I need a living room wall lamp that is in black, and price lower than 100.00 dollars. \\[0.2\baselineskip]
    
    \setlength{\parindent}{0pt} \textbf{AI}:\ \ \ \  search[black living room wall lamp] \\[0.2\baselineskip]
    
    \setlength{\parindent}{0pt} \textbf{Obs}: [Next]

    \setlength{\parindent}{17pt} [A] Generic Raven Decor Crow Lamp Bird Desk Lamp Creative Bedroom Bedside Wall Sconce lamp, \$35. 
    
    \setlength{\parindent}{17pt} [B] Ayux Living Room Dining Room E27 White Deer Head Wall LightGolden Antlers Wall Sconce Rustic Style, \$162.
    
    \setlength{\parindent}{17pt} [C] Dolity Crow desk Lamp, Bird Lamps Bedroom Resin Crow Desk Bedside Light Wall Sconce Light - Black Table lamp. \$33.8 \\[0.2\baselineskip]

    \setlength{\parindent}{0pt} \textbf{AI}:\ \ \ \  The first product is not a wall lamp, the second product is not in black,
    so I should skip them and check the next one, {\color{red} click[C]} \\[0.2\baselineskip]

    \textbf{Obs}: [back to search]
    
    \setlength{\parindent}{17pt} [buy now]

    \setlength{\parindent}{0pt} \textbf{AI}:\ \ \ \  {\color{darkgreen} The third product is a table lamp, not a wall lamp, so I should skip it and check the next one. click[back to search] }\\[0.2\baselineskip] 

    \textbf{Obs}: \(\dots\)

\end{tcolorbox}
\vspace{-10pt} 
\end{wrapfigure}

Moreover, we observed that some negative expert trajectories contain \textbf{recovery} actions, where the expert attempts to recover from harmful expert actions.
Take \textbf{Box 2} as an example. Although GPT-3 correctly analyzed item A and item B, it overlooked the description of item C and mistakenly selected it (click[C]). After this harmful action, the expert attempted to recover by performing a new search using a \textbf{Back} action, which we regard as a recovery action.
In contrast, although our agent can successfully complete this subtask when starting from the beginning, it failed to perform the recovery action when initialized from the expert's mistaken state.
This observation motivates us to incorporate the unlearned recovery segments of these trajectories into training.

Based on these observations, our goal is to include these beneficial expert actions in the RFT fine-tuning process while excluding the remaining expert actions.
For instance, in Fig.~\ref{fig:motivation}, we consider the action sequence \([a_0, \dots, a_{l-1}]\) as beneficial for solving  state \(s_0\), and similarly, \([a_{2l}, \dots, a_{3l-1}]\) as beneficial for solving state \(s_{2l}\) which requires recovery. 
In contrast, the intermediate actions \([a_l, \dots, a_{2l-1}]\) are discarded, as the simulation failure at \(s_{2l}\) suggests that they may not represent useful behavior for the policy to learn.

\subsection{Exploring Expert Failure (EEF)}

\begin{algorithm}[t]
\SetAlgoLined
\DontPrintSemicolon
\caption{Eploring Expert Failure (EEF)}
\label{alg:EEF}
\begin{spacing}{1.3} 
\begin{algorithmic}[1]

\STATE {\bfseries Inputs:} Expert dataset: \(D_e\), Initial policy: \(\pi_{\theta}\)

\STATE {\bfseries Parameters:} Finetune Iteration: $I$, Simulation Num: $M$

\STATE  \(D^+ \leftarrow \{ \tau_e \in D_e : R(\tau_e)=1\}\)  \tcp*{\(D^+\) is a pos trajectory repository}
\STATE Optimize $\theta$ with \(D^+\) and \(\mathcal{L}_{\text{SFT}}\) \tcp*{BC on pos expert trajectories}

\FOR{$i=1, 2, \dots, I-1$}
    \STATE \(D_{i} \leftarrow \{\tau \sim \pi_{\theta}(\cdot \mid s_0) : s_0 \in C_{\text{train}}\}\)  \tcp*{Explore initial states with \(\pi_{\theta}\)}
    \FOR{\(\tau_e=[s_0, s_1, \dots] \in D_e\)} 
        \STATE \(l = \lfloor |\tau_e| / (M+1) \rfloor \) \tcp*{Skip length}
        \STATE \(D_{i}  \leftarrow D_{i} \cup \{\tau \sim \pi_{\theta}(\cdot \mid s_{m\times l}):m \in [1, 2,\dots, M]\}\) \tcp*{Explore states of \(\tau_e\) }
    \ENDFOR
    \STATE \(D^+ \leftarrow D^+ \cup \{\tau \in D_{i} : R(\tau)=1\}\) \tcp*{Add pos trajectories to repository}
    \STATE \(D_{s_0} \leftarrow \{\text{get\_traj}(s_0, D^+) :s_0 \in C_{\text{train}}\}\) \tcp*{Get solutions for all \(s_0\) if exist}
    \STATE \(S_{\text{r}} \leftarrow \text{need\_recover\_states}(D_i)\) \tcp*{Get expert states that \(\pi_{\theta}\) start failing}
    \STATE \(D_{\text{r}} \leftarrow \{\text{get\_traj}(s, D^+) :s \in S_{\text{r}}\}\) \tcp*{Get recovery solutions if exist}
    \STATE Optimize $\theta$ with \(D_{s_0} \cup D_{\text{r}} \) and \(\mathcal{L}_{\text{SFT}}\)
\ENDFOR
\RETURN $\pi_{\theta}$
\end{algorithmic}
\end{spacing}
\end{algorithm}

Based on this motivation, we introduce our method, \textbf{Exploring Expert Failure (EEF)}, outlined in Algo.\ref{alg:EEF}. Similar to previous works such as RFT~\citep{yuan2023scaling} and ETO~\citep{song2024trial}, our method consists of three main phases: (1) \textbf{Behavior Cloning} (line 4), (2) \textbf{Exploration} (lines 6–10), and (3) \textbf{Reinforcement Fine-tuning} (lines 11–15). First, we employ behavior cloning to enhance the capability of a small model. Next, we iteratively perform exploration and reinforcement training phases to further refine the model. In each iteration, the exploration phase simulates all subtasks \(s_0 \in C_{\text{train}}\) and the states selected from the expert trajectories $\tau_e\in D_e$ with the current policy \(\pi_{\theta}\), and the reinforcement training phase analyzes these simulated trajectories, subsequently training the model with actions identified as beneficial.  

The inputs include an expert dataset \(D_e\) and an initial policy $\pi_\theta$. The parameters include the number of reinforcement fine-tuning iterations \(I\), and the number of simulations \(M\) for each expert trajectory per iteration. When the computational budget is limited, one can reduce \(M\) to trade off estimation accuracy for efficiency. The details of each phase are provided in the following paragraphs.







\paragraph{Behavior Cloning Phase}
In this phase, we train our policy to acquire fundamental skills by imitating expert behaviors.
First, like RFT, we reject the negative and select only positive expert trajectories as our training dataset \(D^+\) in Algo~\ref{alg:EEF} line 3. 
Next, we fine-tune our LLM model \(\pi_{\theta}\) with auto-regressive loss in Algo~\ref{alg:EEF} line 4. 
Specifically, given an expert trajectory \(\tau_e = \left(s_0, a_0, s_1, \dots \right),\) we first convert it into a pure text sequence \(\tau = \left(o_0, a_0, o_1, \dots \right),\) that only includes observations and actions. Next, we concatenate the observations $o$ and actions $a$ as a string and convert them to a sequence of tokens \(\mathbf{t} = [t_{0}, t_{1}, \ldots, t_{L}]\), where $L$ is the length of the sequence. 
Since \(\pi_{\theta}\) should only train on the action part of the sequence, for each token \(t_{l}\), we define the loss mask as
\(m_{l}=\mathbbm{1} (\exists\, i \quad \text{s.t.} \quad t_l \in a_i)\), where the indicator \(\mathbbm{1}\) returns 1 if $t_l$ belongs to one of the actions \(a_i\). Finally, the masked autoregressive loss is defined as:
\[
\mathcal{L}_{\text{SFT}}(\pi_{\theta})=-\sum_l m_l \times \log \pi_{\theta}(t_l | t_{<l}).
\]
With \(\mathcal{L}_{\text{SFT}}\), we update the weights \(\theta\) of our policy \(\pi_{\theta}\) in Algo.~\ref{alg:EEF} line 4 to obtain a decent policy to conduct the following exploration.

\paragraph{Exploration Phase}
In the exploration phase, the current model interacts with the environment to collect trajectories and rewards from the environment. Our exploration phase consists of two types of exploration. The first type is the same as RFT, where the policy \(\pi_{\theta}\) explores all subtasks \(s_0 \in C_{\text{train}}\), as presented in line 6 of Algo.~\ref{alg:EEF}.

The second type of exploration (Algo.~\ref{alg:EEF}, lines 7–9) simulates \(M\) expert states for each expert trajectory. 
Our goal is to identify expert actions that are beneficial to our policy, particularly those actions helpful in solving challenging subtasks or recovering from expert mistakes.
The parameter \(M\) controls the computational budget, as some expert trajectories contain an excessive number of states (\(|\tau_e| \gg M\)), making it computationally impractical to simulate all states. 
Therefore, given parameter \(M\), EEF selects expert states at equal intervals. 
Specifically, given an expert trajectory \(\tau_e = [s_0, a_0, \dots]\), EEF first computes a skip length \(l = \lfloor |\tau_e| / (M+1) \rfloor\) and simulates only the selected expert states \([s_{l}, s_{2l}, \dots, s_{M \times l}]\).

In iteration \(i\), all the trajectories generated during the exploration phase are stored in the dataset \( D_i \) for subsequent analysis. Furthermore, all positive trajectories are also stored in the positive trajectory repository \( D^+ \) as solutions for future training.




\paragraph{Reinforcement training phase}
After the exploration phase, we have collected numerous positive trajectories in \(D^+\), either generated from scratch or from expert states. However, fine-tuning our agent on all the trajectories in \(D^+\) is impractical due to high computational costs and may introduce biases toward subtasks with more positive trajectories in \(D^+\)~\cite{Zelikman2022STaRBR}. 
Hence, in this phase, our goal is to determine which trajectories in \(D^+\) are beneficial for training and to identify the specific actions within these trajectories on which the agent should be trained.  
To accomplish this, we adopt a two-step strategy: the \textbf{important state selection} step, in which EEF identifies states considered important, and the \textbf{solution selection} step, where we select a solution from \(D^+\) for each important state, if such solutions exist. The selected solutions are the positive trajectories used for training.
We detail each step in the following paragraphs.



In the \textbf{important state selection} step, 
EEF selects two types of important states.
The first type is the initial states \(s_0\) of each subtask, ensuring that these foundational states remain represented to prevent our agent from forgetting how to solve them. 
The second type consists of states requiring recovery from harmful expert actions. In EEF, we identify these harmful actions through simulations conducted during the exploration phase. Specifically, if the current policy \(\pi_{\theta}\) succeeds when starting from expert state \(s_{i-l}\) but fails when starting from expert state \(s_{i}\), we infer that harmful actions exist within the action sequence \(a_{i-l:i-1}\) and designate \(s_i\) as a state requiring recovery.
To avoid overemphasizing any single expert trajectory, EEF selects only the first expert state requiring recovery from each trajectory as part of the second type of important states.  
Formally, given an expert trajectory \(\tau_e = [s_0, s_1, \dots]\) and \(M\) simulated states \([s_{l}, s_{2l}, \dots, s_{M\times l}]\), alongside corresponding trajectories \([\tau_{s_0}, \tau_{s_l}, \tau_{s_{2l}}, \dots]\) generated by the current policy \(\pi_{\theta}\), we define the state \( s_\text{need\_recover} \) as:
\[
s_\text{need\_recover} = s_{i^*}, \quad \text{where} \quad i^* = \operatorname*{argmin}_{i \in [l, 2l, 3l, \dots, M\times l]} \{ i \mid R(\tau_{s_{i - l}}) = 1, R(\tau_{s_i}) = 0 \}.
\]
As shown in the equation, \( s_{i^*} \) is the first state where the current policy fails (\(R(\tau_{s_{i^*}})=0\)) after previously succeeding (\(R(\tau_{s_{i^*-l}})=1\)), indicating harmful actions within \(a_{i^*-l:i^*-1}\).  
For instance, in Fig.~\ref{fig:motivation}, \( i^* = 2l \) because the agent succeeds at \( s_{l} \) but fails at \( s_{2l} \). This implies expert actions \( [a_l, \dots, a_{2l-1}] \) may transition the agent to overly challenging states. Thus, if recovery actions enabling success from \( s_{2l} \) exist, we aim for the agent to learn these actions, as they may be generalized to other subtasks.


Next is the \textbf{solution selection} step, where we choose at most one solution for each identified important state from \(D^+\) to avoid over-emphasizing any important state. A solution for a given state \( s \) is defined as a positive trajectory containing state \( s \). 
In Algo.~\ref{alg:EEF}, we use the function \(\text{get\_traj}(s, D^+)\) to get the solution path from the positive trajectory repository \(D^+\). 
If there is no solution path in \(D^+\), the function returns None. 
If there is only one solution path in \(D^+\), the function returns the only positive trajectory. 
If multiple solution paths exist for an important state, we choose the solution path with fewer expert actions. 
Training the agent on these solution paths with fewer expert-dependent steps minimizes the necessary gradient adjustments and model updates, preserving learning capacity and allowing the model to focus more on other subtasks or states.
Consider Fig.~\ref{fig:motivation} as an example: state \( s_0 \) has two solutions: \(\tau_{s_l}\) and \(\tau_{s_{3l}}\). In this scenario, we select \(\tau_{s_l}\) as the solution, as it involves fewer expert-generated actions. Thus, the agent only needs to learn the expert actions \([a_0, \dots, a_{l-1}]\) since the subsequent actions after the state \( s_l \) of \(\tau_{s_l}\) are already known to the agent.

After collecting the solution trajectories for the two categories of important states, stored respectively in \(D_{s_0}\) and \(D_{\text{r}}\) as detailed in Algo.~\ref{alg:EEF}, we fine-tune our agent using on \(D_{s_0}\cup D_{\text{r}}\) (Algo.~\ref{alg:EEF}, line 14). In particular, \textbf{when a trajectory is selected as the solution for a specific state  \( s_i\), 
the agent is trained only on the actions \([a_i, a_{i+1}, \dots]\) that occur after state \( s_i \)}
That is, we disable loss propagation for earlier actions \([a_{0}, \dots, a_{i-1}]\) before \(s_i\).
This selective training prevents the agent from incorporating potentially problematic actions preceding state \( s_i \). 
For instance, in Figure~\ref{fig:motivation}, trajectory \(\tau_{s_{3l}}\) provides a solution for state \(s_{2l}\). Although actions \(a_0, a_1, \dots, a_{2l-1}\) are present within \(\tau_{s_{3l}}\), we exclusively train our agent on actions \(a_{2l}, a_{2l+1}, \dots\), as only these subsequent actions are relevant for solving state \(s_{2l}\). By focusing learning solely on actions following important states, our agent reduces the risk of imitating suboptimal expert behaviors.
\begin{wraptable}[11]{R}{0.5\textwidth}
\vspace{-1.em}
    \centering
    \caption{\footnotesize Statistics of datasets. \textbf{Total expert \#}: the number of expert trajectories.  \textbf{Avg Len}: the average length of expert trajectories.}
    \begin{tabular}{lccc}
        \toprule
        & \shortstack{Total\\Expert \#} & \shortstack{Positive\\Expert \#} & \shortstack{Avg\\Len} \\
        \midrule
        Webshop 11k & 11338 & 4106 & 8.26 \\
        Webshop 3k & 2835 & 1045 & 8.24 \\
        Sciworld 2k & 2120 & 1489 & 20.2 \\
        \bottomrule
    \end{tabular}
    \label{tab:expert_annotations}
\end{wraptable}

\section{Experiments}

\subsection{Experimental Settings}
\paragraph{Datasets}
We conducted our experiments in three datasets: WebShop 11k, WebShop 3k, and ScienceWorld 2k (Table~\ref{tab:expert_annotations}). 
Specifically, WebShop 11k~\citep{Ma2024AgentBoardAA} allows us to examine whether our approach can effectively utilize a substantial amount of training data. To investigate generalization under limited data availability, we employ WebShop 3k, a random subset of WebShop 11k containing only 1k positive demonstrations. Additionally, we test our method on ScienceWorld~\citep{Wang2022ScienceWorldIY}, which is designed for scenarios involving longer trajectories (approximately 20 steps). 
Unlike prior works such as ETO, we restrict our agents to only win/loss feedback across all environments and explicitly prohibit teleport actions in SciWorld to increase task difficulty.

\begin{wraptable}{r}{0.5\textwidth}

\centering
\caption{\footnotesize Methods requiring fine-tuning. 
\textbf{Use Neg} indicates if the method uses negative expert trajectories. 
\textbf{FT Iter} is the total iteration number \(I\).
\textbf{Sim. \#} refers to the number of simulations per subtask per iteration. \textbf{Use GPT-3} indicates if GPT-3.5-Turbo-generated trajectories (\(30\times\) cheaper than GPT-4) are also used.}
\vspace{-5pt}
\resizebox{0.5\textwidth}{!}{%
\begin{tabular}{lcccc}
\toprule
 & \shortstack{Use \\ Neg} & \shortstack{FT \\ Iter} & \shortstack{Sim \\ \#} & \shortstack{Use \\ GPT-3}\\
\midrule
SFT ALL       & \yes & 1 & 0 & \no\\[2pt]
SFT POS       & \no  & 1 & 0 & \no\\[2pt]
NAT~\citep{wang2024learning}         & \yes & 1 & 0 & \no\\[2pt]
\midrule
ETO~\citep{song2024trial}          & \no  & 3 & 1 & \no\\[2pt]
RFT~\citep{yuan2023scaling}          & \no  & 3 & 1 & \no\\[2pt]
RFT $\times$6 & \no  & 3 & 6 & \no\\[2pt]
EEF GPT-4     & \yes & 3 & 6 & \no\\[2pt]
EEF GPT-3\&4  & \yes & 3 & 11 & \yes\\
\bottomrule
\end{tabular}

}

\end{wraptable}

For expert demonstrations, we used GPT-4 to generate trajectories based on human instructions within the WebShop datasets~\citep{liu2024agentlite}.
Furthermore, in certain evaluations, GPT-3.5-Turbo-generated demonstrations were used to assess the robustness of our methods under conditions involving weaker expert guidance. 
In contrast, for ScienceWorld, the demonstrations were sourced directly from AgentGym~\citep{xi2024agentgym}. Due to existing models' limitations in achieving high performance, AgentGym supplemented 500 demonstrations from a rule-based gold agent with an additional 1,620 demonstrations produced by GPT-4 Turbo. 


\paragraph{Our method}
We utilize LLAMA3 8B Instruct~\citep{llama3modelcard} as the initial model for fine-tuning. Each fine-tuning iteration consists of six epochs, with a batch size of 64 and a learning rate of 5e-5. All experiments were conducted with four NVIDIA A6000 GPUs. The default parameters are $M=5, I=4$. We select the best model from all iterations. For the Webshop dataset, we evaluate two variants of our method: ``EEF GPT-4'' and ``EEF GPT-3\(\And\)4''. ``EEF GPT-4'' uses the same dataset as the baselines, while ``EEF GPT-3\(\And\)4'' additionally incorporates the data generated by GPT-3 Turbo in the exploration phase and the reinforcement training phase. 


\paragraph{Baselines}
Our baselines can be categorized into three types. The first type includes models \textbf{without fine-tuning}, specifically GPT-3 Turbo and GPT-4. The second type comprises models that are  \textbf{finetuned without exploration}; within this group, we have three baselines. The ``SFT All'' baseline learns from all demonstrations, regardless of their correctness. The ``SFT POS'' baseline learns exclusively from positive demonstrations. 
The Negative-Aware Training (``NAT'')~\citep{wang2024learning} baseline uses the entire dataset but requires the agent to generate incorrect trajectories for questions associated with erroneous demonstrations in the prompt.
The third category of baselines are \textbf{finetuned with exploration}. For this category, we selected RFT and ETO~\citep{song2024trial}. 
We tested two RFT variants: ``RFT,'' which conducts a single exploration per subtask per iteration, and ``RFT\(\times 6\),'' 
which explores each subtask six times, each with different temperature settings ranging from 0.2 to 0.95.
In contrast to RFT's use of SFT, ETO employs DPO~\citep{Rafailov2023DirectPO}, where agent-generated trajectories are compared directly with expert demonstrations based on environment reward (still provided) to produce labeled training pairs.
PPO is excluded due to its significantly lower performance in preliminary experiments~\cite{song2024trial} on unseen subtasks compared to ETO.


\begin{table*}[t]
\centering
\caption{Win rates and rewards of various methods on three agentic environments. Rewards ranges: [0, 1] for WebShop-11k/3k, and [0, 100] for ScienceWorld-2k.}
\begin{tabular}{lccccccc}
\toprule
\multirow{2}{*}{} & \multirow{2}{*}{\parbox{0.8cm}{\centering FT Iter}} & \multicolumn{2}{c}{Webshop 11k} & \multicolumn{2}{c}{Webshop 3k} & \multicolumn{2}{c}{ScienceWorld 2k} \\ \cmidrule(lr){3-4} \cmidrule(lr){5-6} \cmidrule{7-8}
 &  & Winrate & Reward & Winrate & Reward & Winrate & Reward \\
\midrule
 GPT-3.5 Turbo                  & 0 & 23.2\% & 0.60 & 23.2\% & 0.60  & -- & 7.64 \\
 GPT-4                          & 0 & 35.6\% & 0.66 & 35.6\% & 0.66  & -- & 14.4 \\
 \midrule
 SFT ALL                        & 1 & 37.2\% & 0.66 & 39.6\% & 0.68 & 53.0\%  & 68.1 \\
 SFT POS                        & 1 & 46.4\% & 0.75 & 39.6\% & 0.67 & 61.0\%  & 76.8 \\
 NAT                            & 1 & 37.2\% & 0.66 & 40.4\% & 0.68 & 54.0\%  & 69.2 \\
 \midrule
 ETO                            & 3 & 42.0\% & 0.68 & 37.8\% & 0.67 & 57.5\% & 75.0 \\
 RFT                            & 3 & 52.0\% & 0.75 & 38.8\% & 0.66 & 61.5\% & 74.6 \\
 RFT x 6                        & 3 & 53.6\% & 0.76 & 41.4\% & 0.68 & 62.5\% & 73.4 \\
 EEF GPT-4                      & 3 & 58.4\% & 0.78 & 46.8\% & 0.72 & \textbf{68.5\%} &  \textbf{81.3} \\
 EEF GPT-3 \& 4                 & 3 & \textbf{62.0}\% & \textbf{0.81} & \textbf{50.0\% }& \textbf{0.73} & -- & -- \\
\bottomrule
\end{tabular}
\label{tab:main_table}
\end{table*}
\subsection{Main Results}
Table~\ref{tab:main_table} presents the performance comparison between our proposed EEF method and other baselines in three datasets.  
The results show a clear performance advantage of our proposed method compared to other baselines, including the GPT-4 baseline, across all evaluated datasets. Specifically, the EEF GPT-4 variant achieves significant improvements over GPT-4, increasing the win rate from 35.6\% to 58.4\% on Webshop 11k, from 35.6\% to 46.8\% on Webshop 3k, and notably from lower than 14.4\% to 68.5\% on ScienceWorld 2k. 
Furthermore, compared to the best-performing fine-tuning baseline (RFT \(\times\) 6), EEF GPT-4 demonstrates additional improvements, raising the win rate from 53.6\% to 58.4\% on Webshop 11k, from 41.4\% to 46.8\% on Webshop 3k, and from 62.5\% to 68.5\% on ScienceWorld 2k. Additionally, when incorporating data from GPT-3 Turbo into EEF (EEF GPT-3 \& 4), we observe even greater performance gains, particularly on the Webshop datasets, where the win rate further rises to 62.0\% and 50.0\% for Webshop 11k and Webshop 3k, respectively. 
This highlights that our method effectively leverages additional demonstrations from a weaker expert (GPT-3.5 Turbo), achieving notable performance improvements with only a slight increase in cost.


\subsection{Ablation Studies}
\begin{wrapfigure}[13]{R}{0.4\textwidth}
\vspace{-1em}
  \centering
  \includegraphics[width=0.4\textwidth]{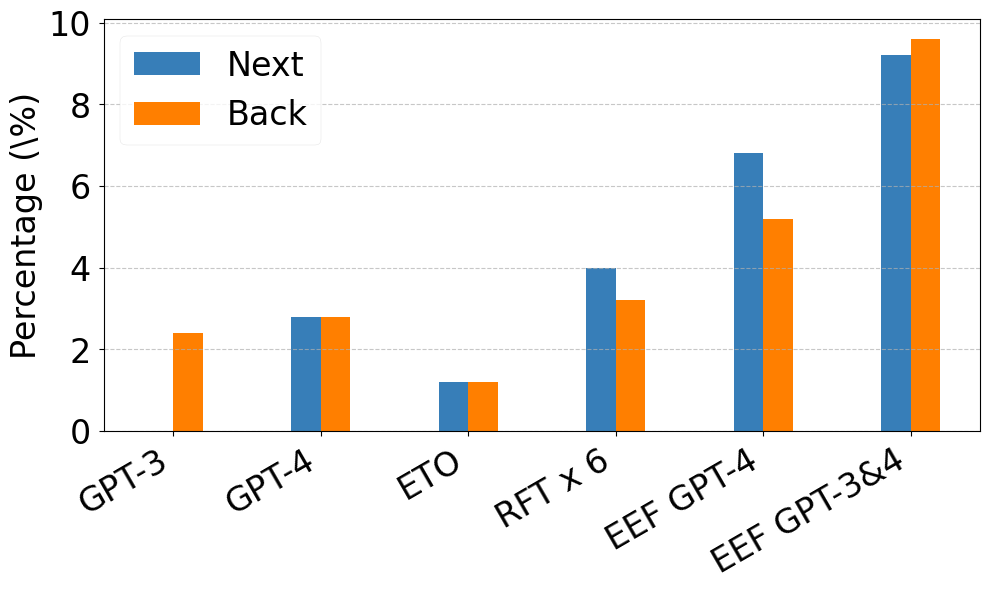}
  \vspace{-1.5em}
  \caption{The percentage of successful subtasks that involve the use of navigation skills: \textbf{Next} and \textbf{Back}.}
  \label{fig:next}
\end{wrapfigure}
\paragraph{Navigation skills}
In this experiment, we evaluate the mitigation of simplicity bias by measuring the effectiveness of navigation skills on the Webshop 11k dataset. Specifically, we focus on two types of navigation actions: ``Next page'' (Next) and ``Back to search'' (Back), as shown in Fig.~\ref{fig:next}.
Our methods, EEF GPT-4 and EEF GPT-3\&4, exhibit superior proficiency in utilizing these navigation skills compared to baseline models. 
In addition, we investigated the attempt rate of using navigation actions. The results are shown in Table~\ref{tab:next} of Appendix~\ref{sec:nav}.
In particular, while GPT-3.5 and GPT-4 display a lower percentage of successful subtasks that involve navigation actions, they actually show relatively high attempt rates. 
For example, GPT-4 attempted to use the Next action in 16.8\% of the subtasks. This indicates that GPT-4 recognizes the potential usefulness of such navigation actions but lacks the ability to execute them effectively. In contrast, our method not only identify these beneficial actions in the negative expert trajectories but also convert them into successful task completions through more effective skill utilization.



\begin{wrapfigure}{RT}{0.4\textwidth}
\vspace{-10pt}
  \centering
  \includegraphics[width=0.4\textwidth]{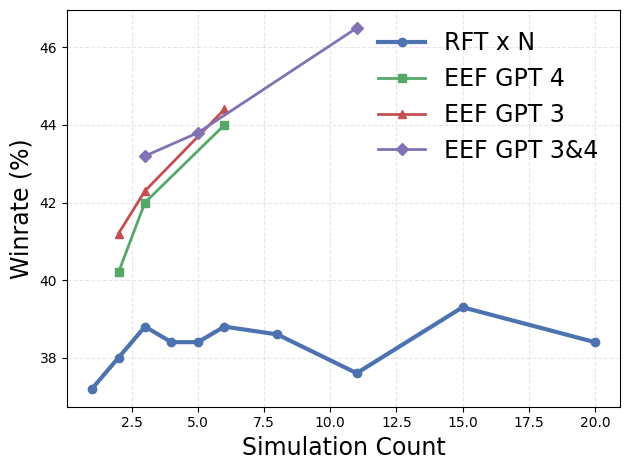}
  \caption{
Win rates of different methods training the same base model for one iteration under varying simulation budgets (exploration cost) by adjusting $N$ or $M$. \textbf{RFT × N}: Initial state \(s_0\) simulated \(N\) times. \textbf{EEF}: Different expert state simulation numbers \(M=1,2,5\) on different expert datasets. }
  \label{fig:efficiency}
\vspace{-5pt}
\end{wrapfigure}
\paragraph{Case studies} 
To further substantiate the necessity of navigation skills for challenging subtasks, we examine specific subtasks where (1) our method succeeded by effectively employing these skills, (2) RFT methods failed due to the absence of these skills, and (3) GPT-4 applied these skills but ultimately failed.
Some representative samples of these subtasks, along with a side-by-side comparison of trajectories generated by the RFT method and our method, can be found in Appendix~\ref{AP: next case studies} and ~\ref{AP: back case studies}.
Upon examining the trajectores of these type subtasks, it becomes evident that navigation skills are crucial. 
For subtasks that require the Next action, the products initially presented do not satisfy user requirements. For instance, listed products may have incorrect attributes such as an inappropriate price range or incorrect colors (e.g., ask for petal green but get petal blue).
Similarly, for subtasks that require the Back action, the items initially selected appear suitable on the search page, but ultimately lack specific attributes such as desired sizes or colors upon closer inspection on the product page. (e.g., ask for a 250ml shampoo but only find 500ml or 200ml options).
Another notably challenging subtask involved a user's request for a high-power sound system. Here, the agent needed to recognize that a portable 16W soundbar was insufficient and consequently navigate to the next page to identify a more suitable product.
These case studies clearly demonstrate that our approach significantly mitigates simplicity bias and is able to employ advanced navigation skills.

\paragraph{Efficiency Analysis}
In this section, we evaluate the exploration efficiency of our proposed method by varying the number of simulations, specifically setting \(M=1,2,5\).
We compared these results against a baseline exploration method that utilizes different temperatures within the Webshop 3k environment.
To ensure fairness in our comparison, we employed the same model (SFT POS) for conducting explorations, and subsequently utilized the resulting data to finetune a new model with only one iteration. 
The results presented in Fig.~\ref{fig:efficiency} clearly demonstrate the superior performance of our method relative to the baseline exploration approach. Specifically, the baseline method fails to achieve a 40\% winrate even when employing up to 11 simulations. In contrast, our method achieves a 40\% winrate using merely 2 simulations (when \(M=1\)) and consistently improves performance with increased simulation budgets.
Furthermore, our findings indicate that exploring the trajectories generated by GPT-3 consistently yields better results than exploring those generated by GPT-4. This phenomenon may be attributed to the exploration model being trained exclusively on positive GPT-4 data. Therefore, adding the GPT-3.5 Turbo trajectories can offer richer learning opportunities.

\paragraph{Different Models} To further validate the effectiveness and generalization of our method, we conducted additional experiments using one of the latest models, mistral-7b-v0.3, on the Webshop 3k dataset. The results demonstrate that our approach consistently outperforms existing baselines. Specifically, our method achieved win rates of 46.8\% and 48.8\% with the ``GPT-4'' and ``GPT-3\&4'' models, respectively, compared to only 42.2\% obtained by RFTx6. These findings underscore the robust generalization capability of our approach across different initial LLMs.


\section{Related Work}
Recent advances in large language models (LLMs) have prompted researchers to build LLM-based agents for multi-step tasks, exploiting the models’ emergent abilities~\citep{achiam2023gpt,wei2022emergent}.
Broadly, existing approaches can be categorized into methods without fine-tuning (zero-shot or few-shot), methods with fine-tuning but no exploration, and methods with exploration-based fine-tuning. Zero-shot and few-shot methods typically focus on prompt engineering to reduce errors and hallucinations~\citep{huang2022language,huang2022inner,yao2023react,wang2023voyager}.
Approaches that use fine-tuning without exploration
often rely on filtering suboptimal expert trajectories based on rewards~\citep{zeng2023agenttuning,chen2023fireact,chen2024agent},
or adding a negative label on the prompt of negative demonstrations~\citep{wang2024learning}.
and some also train on a large set of tasks to learn general concepts transferable to tasks with limited demonstrations~\citep{zhang2024agentohana, song2024agentbank}.
Exploration-based fine-tuning adds newly discovered positive samples to the supervised fine-tuning dataset~\citep{aksitov2023rest, xi2024agentgym};
experiments have shown that training on self-generated data can be more sample-efficient~\citep{setlur2024rl}.
Other variants (e.g., ETO) use DPO to reduce the likelihood of generating negative trajectories~\citep{song2024trial}. Note that ETO, as well as many methods~\citep{xiong2024watch} that rely on ETO's dataset, use only high-reward expert trajectories.

However, many of these approaches treat the entire trajectory with a single final reward, which can be simplistic. More recent methods adopt a stepwise analysis of trajectories for better finetuing~\citep{wang2023math,wang2024multi,ma2023let,havrilla2024glore,xiong2024watch}.
For example, by training a model to identify critical steps or highly rewarded and then focus training only on those steps~\citep{chen2025atlas, wang2025steca}.
Another strategy assumes that the expert is always correct, applying DPO at each step~\citep{deng2024novice}; 
but this is not viable when the expert frequently fails. Stepwise DPO~\citep{lai2024step}
assumes that if a state contains both positive and negative trajectories, the agent should learn from the positive action. Yet implementing stepwise DPO on the negative expert trajectories still differs from our approach. 
Specifically, given a state \(s_i\) and two trajectories \(\tau_{s_i}^+\) and \(\tau_{s_i}^-\) labeled as chosen and rejected, stepwise DPO increases the probability of actions after \(s_i\) in \(\tau_{s_i}^+\), while decreasing it for \(\tau_{s_i}^-\).  
In contrast, our method emphasizes beneficial actions \emph{before} \(s_i\) that contribute to successful simulations.  
Additionally, stepwise DPO may produce an excessive number of trajectory pairs when multiple states each yield a chosen–rejected pair, whereas our method limits representation to one per subtask and per expert trajectory.

\section{Conclusion and Future Work}
In this paper, we present EEF, a novel framework that learns beneficial actions from negative expert data while remaining robust against noise from suboptimal actions. Remarkably, our method can even learn effectively from trajectories generated by weaker experts such as GPT-3.5. Experimental results demonstrate that EEF achieves state-of-the-art (SOTA) performance on both the Webshop and ScienceWorld environments. Furthermore, EEF retains the simplicity of RFT by relying solely on supervised fine-tuning (SFT) loss, without requiring additional reward model training, making it easier to apply and reducing the need for extensive hyperparameter tuning.

There are several promising directions to further improve our method. For instance, one can perform preference learning by using only beneficial actions as the chosen data. 
Another direction is to enhance the accuracy of identifying these beneficial actions. Instead of searching over a fixed interval, a binary search strategy could be employed to more efficiently determine which actions are advantageous to the model. Furthermore, if computational resources permit, the method could be enhanced by incorporating expert trajectories as the main branches in a tree search algorithm—such as Monte Carlo Tree Search (MCTS)~\citep{coulom2006efficient}—to better solve out-of-distribution (OOD) subtasks or states. 
In addition, the solution selection step can be modified to select the best solution.

Moreover, our results suggest that even when GPT-4 trajectories are available, GPT-3 trajectories still contain valuable information. This opens up opportunities for further research on combining expert datasets, particularly when one expert source is significantly more cost-effective than the other. How to allocate resources between high-quality and low-cost experts becomes an even more critical trade-off challenge.

\bibliography{colm2025_conference}
\bibliographystyle{colm2025_conference}

\newpage
\appendix
\section{Appendix}

\subsection{Navigation actions' success and attempted rate}
\label{sec:nav}

\begin{table}[ht]
\centering
\caption{Performance comparison in terms of pos and all metrics.  Columns 2-3 indicate the proportion of test subtasks successfully solved using Next or Back, respectively. Columns 4-5 show the proportion of attempts made by agents using these navigation skills. }
\label{tab:next}
\begin{tabular}{lrrrr}
\toprule
 & \multicolumn{2}{c}{Success (\%)} & \multicolumn{2}{c}{Attempt (\%)} \\
\cmidrule(lr){2-3} \cmidrule(lr){4-5}
 & Next & Back & Next & Back \\
\midrule
GPT-3.5 Turbo   & 0.0 & 2.4 & 3.2  & 16.8 \\
GPT-4           & 2.8 & 2.8 & 16.8 & 14.0 \\
SFT ALL         & 2.8 & 1.6 & 16.8 & 9.6  \\
SFT POS         & 4.4 & 2.0 & 13.2 & 4.0  \\
ETO             & 1.2 & 1.2 & 9.6  & 3.2  \\
RFT             & 4.0 & 3.2 & 9.6  & 8.4  \\
RFT × 6         & 4.0 & 3.2 & 9.2  & 7.6  \\
EEF GPT-4       & 6.8 & 5.2 & 13.2 & 12.0 \\
EEF GPT-3\&4    & 9.2 & 9.6 & 13.2 & 14.0 \\
\bottomrule
\end{tabular}
\end{table}

\subsection{Case studies of Next action}
\label{AP: next case studies}
\begin{tcolorbox}[colback=blue!5!white, breakable, colframe=blue!75!black, title={Next Example 1: RFT, reward =  0.66}]
\scriptsize
Task:i am looking for resilient memory foam loveseat sofa \\

Action:Think[{"response": "I should first use Search to search for a resilient memory foam loveseat sofa."}] \\

Observation: OK \\

Action:search[{"product": "resilient memory foam loveseat sofa"}] \\

Observation: WEB PAGE: {

[Back to Search] 

Page 1 (Total results: 15) 

[Next] 

[B08737G8HG] 
Modern Fabric Double Sofa Couch Loveseat Linen Accent Chair Comfy Upholstered Arm Lounge Chair for Bedroom Living Room \$279.99 

[B09NS47YJ7] 
Sectional Couches for Living Room Faux Leather Office Couch Sets Mid Century Modern Small Sofa for Bedroom Mini Settee Loveseat Comfy for Small Apartment(Loveseat, Brown) 
\$512.0 

[B09F5J244F] 
Hydeline Erindale Power Recline, Headrest, Lumbar, Top Grain Leather Sofa Couch with Built in USB Ports, 88'', Cognac 
\$2699.0 } \\

Action:Think[{"response": "The first product seems to be the most relevant to the requirements. I should click the first product."}] \\

Observation: OK \\ 

Action:click[{"button": "B08737G8HG"}] \\

Observation: WEB PAGE: {

[Back to Search]  

[Prev] 
Modern Fabric Double Sofa Couch Loveseat Linen Accent Chair Comfy Upholstered Arm Lounge Chair for Bedroom Living Room 
Price: \$279.99 
 } \\ 

Action:click[{"button": "Buy Now"}] \\

Observation: shopping is finished. \\

\end{tcolorbox}

\begin{tcolorbox}[colback=blue!5!white, breakable, colframe=blue!75!black, title={Next Example 1: EEF, reward =  1.00}]
\scriptsize

Task:i am looking for resilient memory foam loveseat sofa \\

Action:Think[{"response": "I should first use Search to search for resilient memory foam loveseat sofa."}] \\

Observation: OK \\

Action:search[{"product": "resilient memory foam loveseat sofa"}] \\

Observation: WEB PAGE: {

[Back to Search] 

Page 1 (Total results: 15) 

[Next] 

[B08737G8HG] 
Modern Fabric Double Sofa Couch Loveseat Linen Accent Chair Comfy Upholstered Arm Lounge Chair for Bedroom Living Room 
\$279.99 

[B09NS47YJ7] 
Sectional Couches for Living Room Faux Leather Office Couch Sets Mid Century Modern Small Sofa for Bedroom Mini Settee Loveseat Comfy for Small Apartment(Loveseat, Brown) 
\$512.0 

[B09F5J244F] 
Hydeline Erindale Power Recline, Headrest, Lumbar, Top Grain Leather Sofa Couch with Built in USB Ports, 88'', Cognac 
\$2699.0 } \\

Action:Think[{"response": "None of the products listed are resilient memory foam loveseat sofas. I should click Next to see more options."}] \\

Observation: OK \\

Action:click[{"button": "Next"}] \\

Observation: WEB PAGE: {

[Back to Search] 

Page 2 (Total results: 15) 

[Prev] 

[Next] 

[B09NS26T9M] 
Faux Leather Office Couch Sets Mid Century Modern Sectional Small Sofa for Bedroom Mini Settee Loveseat Comfy for Living Room Small Apartment(3-Seat Sofa, Black) 
\$512.0 

[B09QMF83B5] 
56-inch Small Loveseat Sofa, Modern Couch with Removable Cushion, Upholstered Living Room Sofa with Rose Golden Metal Legs, Leopard Teal 
\$259.99 

[B0872G9J57] 
Poppy Sofa - Nelson Sectional/Sofa/Loveseat/Loveseat Sectional/Armchair 42 Colors - Resilient Memory Foam - Made in US - Mid Century Modern (Sand Cross Weave Upholstery) SF031720036E1 
\$1751.0 } \\

Action:Think[{"response": "The third product is a resilient memory foam loveseat sofa. I should click on the third product."}] \\

Observation: OK \\

Action:click[{"button": "B0872G9J57"}] \\

Observation: WEB PAGE: { 

[Back to Search]  

[Prev]

size [armchair][loveseat][sectional][sectional loveseat][sofa]

Price: \$1751.0 

[Buy Now] } \\

Action:click[{"button": "loveseat"}] \\

Observation: WEB PAGE: {You have clicked loveseat.} \\

Action:click[{"button": "Buy Now"}] \\

Observation: shopping is finished. \\

\end{tcolorbox}

\begin{tcolorbox}[colback=blue!5!white, colframe=blue!75!black, breakable, title={Next Example 2: RFT, reward =  0.66}]
\scriptsize

Task:i need a detangler hair brush that stimulates hair growth. choose the purple one, and price lower than 50.00 dollars \\

Action:Think[{"response": "I should first use Search to search for a detangler hair brush that stimulates hair growth."}] \\

Observation: OK \\

Action:search[{"product": "detangler hair brush that stimulates hair growth"}] \\

Observation: WEB PAGE: { 

[Back to Search] 

Page 1 (Total results: 15) 

[Next] 

[B08KYNCDVG] 
Zolf Natural Bristle Detangling for Wet/Dry Hair, for Women, Men, and Kids. Includes brush cleaner. 
\$19.77 

[B07QW12YR9] 
Hair Brush, Natural Boar Bristle Hair Brush, Wooden Bamboo Hair Brush for Women Mens, Paddle Brush Curly Hair Brush for Thick Hair \& Anti Static Detangling Best Paddle Brush for Reducing Hair Breakage. 
\$9.99

[B09189QFJ9] 
Hair Brush, Natural Boar Bristle Hair Brush, Wooden Bamboo Hair Brush for Women Mens, Paddle Brush Curly Hair Brush for Thick Hair \& Anti Static Detangling Best Paddle Brush for Reducing Hair Breakage 
\$9.99 } \\

Action:Think[{"response": "The first product seems to be the most relevant to the requirements. I should click the first product."}] \\

Observation: OK \\

Action:click[{"button": "B08KYNCDVG"}] \\

Observation: WEB PAGE: {

[Back to Search] 

[Prev] 

Zolf Natural Bristle Detangling for Wet/Dry Hair, for Women, Men, and Kids. Includes brush cleaner.

Price: \$19.77 
} \\

Action:click[{"button": "Buy Now"}]

\end{tcolorbox}

\begin{tcolorbox}[colback=blue!5!white, breakable, colframe=blue!75!black, title={Next Example 2: EEF, reward =  1.0}]
\scriptsize
Task:i need a detangler hair brush that stimulates hair growth. choose the purple one, and price lower than 50.00 dollars \\

Action:Think[{"response": "I should first use Search to search for a detangler hair brush that stimulates hair growth."}] \\

Observation: OK \\

Action:search[{"product": "detangler hair brush that stimulates hair growth"}] \\

Observation: WEB PAGE: {

[Back to Search] 

Page 1 (Total results: 15) 

[Next] 

[B08KYNCDVG] 
Zolf Natural Bristle Detangling for Wet/Dry Hair, for Women, Men, and Kids. Includes brush cleaner. 
\$19.77 

[B07QW12YR9] 
Hair Brush, Natural Boar Bristle Hair Brush, Wooden Bamboo Hair Brush for Women Mens, Paddle Brush Curly Hair Brush for Thick Hair \& Anti Static Detangling Best Paddle Brush for Reducing Hair Breakage. 
\$9.99 

[B09189QFJ9] 
Hair Brush, Natural Boar Bristle Hair Brush, Wooden Bamboo Hair Brush for Women Mens, Paddle Brush Curly Hair Brush for Thick Hair \& Anti Static Detangling Best Paddle Brush for Reducing Hair Breakage 
\$9.99 } \\

Action:Think[{"response": "None of the products listed stimulate hair growth or are purple. I should click Next to see more options."}] \\

Observation: OK \\

Action:click[{"button": "Next"}] \\

Observation: WEB PAGE: { 

[Back to Search] 

Page 2 (Total results: 15) 

[Prev] 

[Next] 

[B07WN2DJSL] 
Curved Vented Boar Bristle Styling Hair Brush, Single White Brush Anti-static Detangler, Wet or Dry Use, Fast Blow Drying, Use on Long or Short Hair. (1 Pack, White) 
\$7.99 

[B07VQDM2CF] 
Oleh-Oleh Wet and Dry Brush Pro Detangle Hair Brush + Folding Mirror Mini Pop Up Hairbrush (BLUE+PURPLE). Healthy to scalp.
\$16.99

[B00S1L6590] 
Giorgio GION2R Neon Red 6.25 inch Gentle Touch Detangler Hair Brush for Men Women \& Kids. Soft Bristles for Sensitive Scalp. Wet and Dry for all Hair Types. Scalp Massager Brush Stimulate Hair Growth 
\$14.95 } \\

Action:Think[{"response": "The second product is purple and stimulates hair growth. I should click on the second product."}] \\

Observation: OK \\

Action:click[{"button": "B07VQDM2CF"}] \\

Observation: WEB PAGE: {

[Back to Search] 

[Prev] 

color [blue+purple][blue+red+light blue][purple][purple+blue][rose+blue+purple][set1]

Oleh-Oleh Wet and Dry Brush Pro Detangle Hair Brush + Folding Mirror Mini Pop Up Hairbrush (BLUE+PURPLE) 

Price: \$16.99 

[Buy Now] } \\

Action:click[{"button": "purple"}] \\

Observation: WEB PAGE: {You have clicked purple.} \\

Action:click[{"button": "Buy Now"}] \\

Observation: shopping is finished. \\

\end{tcolorbox}
\subsection{Case studies of Back action}
\label{AP: back case studies}
\begin{tcolorbox}[colback=blue!5!white, breakable, colframe=blue!75!black, title={Back Example 1: RFT, reward =  0.66}]
\scriptsize
Task:i want a red office chair ergonomic gaming chair with lumbar support, and price lower than 140.00 dollars \\

Action:Think[{"response": "I should first use Search to search for a red office chair ergonomic gaming chair with lumbar support."}] \\

Observation: OK \\

Action:search[{"product": "red office chair ergonomic gaming chair with lumbar support"}] \\

Observation: WEB PAGE: {

[Back to Search] 

Page 1 (Total results: 15) 

[Next] 

[B098DG5F8T] 
Tyyps Ergonomic Gaming Chair -250lb High Back PC Racing Desk Chair Adjustable Reclining Computer Chair with Lumbar Support Armrest Headrest Task Rolling Swivel Chair Game Chair for Adult Teen, Black 
\$52.95

[B08PPGC1C4] 
Gaming Chair Office Chair Desk Chair Ergonomic High Back Computer Chair with Lumbar Support Flip-up Arms Headrest PU Leather Swivel Task Chair for Women Men Adult, Pink 
\$105.99

[B097BG6P5Q] 
Executive PU Leather Office Chair, Ergonomic High Back Managerial Chairs with Lumbar Support Armrest Swivel, Adjustable Computer Gaming Chair for Office Home Gaming Room Red 
\$162.99 } \\

Action:Think[{"response": "The third product is a red office chair with lumbar support, but it's over the budget. I should click the first product."}] \\

Observation: OK \\

Action:click[{"button": "B098DG5F8T"}] \\

Observation: WEB PAGE: {

[Back to Search] 

[Prev] 

Tyyps Ergonomic Gaming Chair -250lb High Back PC Racing Desk Chair Adjustable Reclining Computer Chair with Lumbar Support Armrest Headrest Task Rolling Swivel Chair Game Chair for Adult Teen, Black 

Price: \$52.95 

[Buy Now] } \\

Action:click[{"button": "Buy Now"}] \\

Observation: shopping is finished. \\

\end{tcolorbox}

\begin{tcolorbox}[colback=blue!5!white, breakable, colframe=blue!75!black, title={Back Example 1: EEF, reward =  1.0}]
\scriptsize
Task:i want a red office chair ergonomic gaming chair with lumbar support, and price lower than 140.00 dollars \\

Action:Think[{"response": "I should first use Search to search for a red office chair ergonomic gaming chair with lumbar support."}] \\

Observation: OK \\

Action:search[{"product": "red office chair ergonomic gaming chair with lumbar support"}] \\ 

Observation: WEB PAGE: {

[Back to Search] 

Page 1 (Total results: 15) 

[Next] 

[B098DG5F8T] 
Tyyps Ergonomic Gaming Chair -250lb High Back PC Racing Desk Chair Adjustable Reclining Computer Chair with Lumbar Support Armrest Headrest Task Rolling Swivel Chair Game Chair for Adult Teen, Black 
\$52.95

[B08PPGC1C4] 
Gaming Chair Office Chair Desk Chair Ergonomic High Back Computer Chair with Lumbar Support Flip-up Arms Headrest PU Leather Swivel Task Chair for Women Men Adult, Pink 
\$105.99

[B097BG6P5Q] 
Executive PU Leather Office Chair, Ergonomic High Back Managerial Chairs with Lumbar Support Armrest Swivel, Adjustable Computer Gaming Chair for Office Home Gaming Room Red 
\$162.99 } \\ 

Action:Think[{"response": "The third product is a red office chair with lumbar support, but it's over the budget. I should click the first product to see if it has a red color option."}] \\

Observation: OK \\

Action:click[{"button": "B098DG5F8T"}] \\

Observation: WEB PAGE: {

[Back to Search] 

[Prev]

Tyyps Ergonomic Gaming Chair -250lb High Back PC Racing Desk Chair Adjustable Reclining Computer Chair with Lumbar Support Armrest Headrest Task Rolling Swivel Chair Game Chair for Adult Teen, Black 

Price: \$52.95

[Buy Now] }\\

Action:Think[{"response": "The product does not have a red color option. I should go back to the search results and try the next product."}] \\

Observation: OK \\

Action:click[{"button": "Back to Search"}] \\

Observation: WEB PAGE: {Let us begin shopping 
[Search]}  \\

Action:search[{"product": "red office chair ergonomic gaming chair with lumbar support"}] \\

Observation: WEB PAGE: { \\

[Back to Search] 

Page 1 (Total results: 15) 

[Next]

[B098DG5F8T] 
Tyyps Ergonomic Gaming Chair -250lb High Back PC Racing Desk Chair Adjustable Reclining Computer Chair with Lumbar Support Armrest Headrest Task Rolling Swivel Chair Game Chair for Adult Teen, Black 
\$52.95 

[B08PPGC1C4] 
Gaming Chair Office Chair Desk Chair Ergonomic High Back Computer Chair with Lumbar Support Flip-up Arms Headrest PU Leather Swivel Task Chair for Women Men Adult, Pink 
\$105.99

[B097BG6P5Q] 
Executive PU Leather Office Chair, Ergonomic High Back Managerial Chairs with Lumbar Support Armrest Swivel, Adjustable Computer Gaming Chair for Office Home Gaming Room Red 
\$162.99 } \\

Action:click[{"button": "B08PPGC1C4"}] \\

Observation: WEB PAGE: {

[Back to Search] 

[Prev] 

color [blue][camo][grey][pink][red][white]
Gaming Chair Office Chair Desk Chair Ergonomic High Back Computer Chair with Lumbar Support Flip-up Arms Headrest PU Leather Swivel Task Chair for Women Men Adult, Pink 
Price: \$105.99 
[Buy Now] } \\

Action:click[{"button": "red"}] \\

Observation: WEB PAGE: {You have clicked red.} \\

Action:click[{"button": "Buy Now"}] \\

Observation: shopping is finished. \\

\end{tcolorbox}

\begin{tcolorbox}[colback=blue!5!white, breakable, colframe=blue!75!black, title={Back Example 2: RFT, reward =  0.75}]
\scriptsize
Task:i am looking for 2 pack of 20ft long quadshield solid copper black color indoor and outdoor coaxial cable, and price lower than 60.00 dollars\\ 

Action:Think[{"response": "I should first use Search to search for 2 pack of 20ft long quadshield solid copper black color indoor and outdoor coaxial cable."}] \\

Observation: OK \\

Action:search[{"product": "2 pack of 20ft long quadshield solid copper black color indoor and outdoor coaxial cable"}] \\

Observation: WEB PAGE: {

[Back to Search] 

Page 1 (Total results: 15) 

[Next] 

[B074741B7G] 
PHAT SATELLITE INTL 100ft White Solid Copper COAXIAL RG6 3GHz UV Jacket Indoor Outdoor Satellite TV Antenna Cable 18AWG 75 Ohm Anti-Corrosion Nickel Compression Connectors Assembled in USA 
\$18.0 

[B07CQ96G4F] 
150ft Black TRI-Shield Indoor Outdoor RG-6 Coaxial Cable Nickel-Plated Brass Connector 75 Ohm (Satellite TV, Broadband Internet, Ham Radio, OTA HD Antenna Coax) Assembled in USA PHAT SATELLITE 
\$19.5

[B07CQB2FZK] 
200ft Black TRI-Shield Indoor Outdoor RG-6 Coaxial Cable Nickel-Plated Brass Connector 75 Ohm (Satellite TV, Broadband Internet, Ham Radio, OTA HD Antenna Coax) Assembled in USA PHAT SATELLITE 
\$21.5 } \\

Action:Think[{"response": "None of the products match the exact search criteria. I should click on the first product as it is the closest match."}] \\

Observation: OK \\

Action:click[{"button": "B074741B7G"}] \\
 
Observation: WEB PAGE: {

[Back to Search] 

[Prev]

size [1ft][1ft (3 pack)][2ft][2ft (3 pack)][3 ft][3ft][3ft (3 pack)][4ft][4ft (3 pack)][5 ft][5ft][5ft (3 pack)][6 ft][6ft][6ft (3 pack)][8ft][8ft (3 pack)][10 ft][10ft][10ft (3 pack)][12 ft][12ft][12ft (3 pack)][15 ft][15ft][15ft (2 pack)][20 ft][20ft][20ft (2 pack)][25 ft][25ft][25ft (2 pack)][30 ft][30ft][30ft (2 pack)][35ft][35ft (2 pack)][40 ft][40ft][40ft (2 pack)][45ft][45ft (2 pack)][50 ft][50ft][50ft (2 pack)][55ft][55ft (2 pack)][60 ft][60ft][60ft (2 pack)][65ft][65ft (2 pack)][70ft][70ft (2 pack)][75ft][75ft (2 pack)][80ft][85ft][90ft][95ft][100 ft][100ft][105ft][110ft][115ft][120ft][125ft][130ft][135ft][140ft][145ft][150 ft][150ft][155ft][160ft][165ft][170ft][175ft][180ft][185ft][190ft][195ft][200 ft][200ft]

color [plenum cmp catvp - white][solid copper - black][solid copper w |  weather boot - white][bare copper w |  digicon connector - white][direct burial tri-shield - black][direct burial tri-shield, boot - black][quad-shield w |  weather boot - black][quadshield - black][quadshield - white][quadshield solid copper - black][solid copper - white]

PHAT SATELLITE INTL 100ft White Solid Copper COAXIAL RG6 3GHz UV Jacket Indoor Outdoor Satellite TV Antenna Cable 18AWG 75 Ohm Anti-Corrosion Nickel Compression Connectors Assembled in USA 

Price: \$18.0 
[Buy Now] } \\

Action:click[{"button": "20ft (2 pack)"}] \\

Observation: WEB PAGE: {You have clicked 20ft (2 pack).} \\

Action:click[{"button": "solid copper - black"}] \\

Observation: WEB PAGE: {You have clicked solid copper - black.} \\

Action:click[{"button": "Buy Now"}] \\

Observation: shopping is finished. \\

\end{tcolorbox}

\begin{tcolorbox}[colback=blue!5!white, breakable, colframe=blue!75!black, title={Back Example 2: EEF, reward =  1.0}]
\scriptsize

Task:i am looking for 2 pack of 20ft long quadshield solid copper black color indoor and outdoor coaxial cable, and price lower than 60.00 dollars \\

Action:Think[{"response": "I should use the Search function to find the coaxial cable with the specific requirements."}] \\

Observation: OK \\

Action:search[{"product": "20ft long quadshield solid copper black color indoor and outdoor coaxial cable"}] \\

Observation: WEB PAGE: {

[Back to Search] 

Page 1 (Total results: 15) 

[Next] 

[B073L44ZBG] 
20ft White TRI-Shield Weather Seal Indoor Outdoor RG-6 Coaxial Cable Brass Connector 75 Ohm Satellite TV Broadband Internet Ham Radio OTA HD Antenna Coax Assembled in USA by PHAT SATELLITE INTL 
\$11.4 

[B074741B7G] 
PHAT SATELLITE INTL 100ft White Solid Copper COAXIAL RG6 3GHz UV Jacket Indoor Outdoor Satellite TV Antenna Cable 18AWG 75 Ohm Anti-Corrosion Nickel Compression Connectors Assembled in USA 
\$18.0 

[B073L2RDKV] 
PHAT SATELLITE INTL Tri-Shielded Solid Core 18AWG RG-6 Coaxial Cable, Satellite TV Broadband Modem Internet OTA HD Antenna Indoor Outdoor Coax Applications, UL ETL, Assembled in USA, (200 feet White) 
\$28.5 }\\

Action:Think[{"response": "None of the products match the specific requirements. I should search again with different keywords."}] \\

Observation: OK \\

Action:search[{"product": "quadshield solid copper black coaxial cable"}] \\

Observation: WEB PAGE: {There is no [Search] button, you should click the [Back to Search] button first to search something } \\

Action:click[{"button": "Back to Search"}] \\ 

Observation: WEB PAGE: {Let us begin shopping 
[Search]} \\

Action:search[{"product": "quadshield solid copper black coaxial cable"}] \\

Observation: WEB PAGE: {

[Back to Search] 

Page 1 (Total results: 15) 

[Next ] 

[B07H9G2RPM] 
PHAT SATELLITE INTL - 3GHz Direct Burial Underground RG6 Coaxial Cable, Tri-Shield Coax, Gel Coated Braids, Weather Boot Compression Connectors, Satellite Approved, Made in USA (12 feet, Orange) \$13.0 

[B07H9GYZPD] 
PHAT SATELLITE INTL - 3GHz Direct Burial Underground RG6 Coaxial Cable, Tri-Shield Coax, Gel Coated Braids, Weather Boot Compression Connectors, Satellite Approved, Made in USA (100 feet, Orange) 
\$30.5 

[B073KR5D4W] 
135ft Black Quad Shield Solid Copper 3GHZ RG-6 Coaxial Cable 75 Ohm (DIRECTV Satellite TV or Broadband Internet) Anti Corrosion Brass Connector RG6 Fittings Assembled in USA by PHAT SATELLITE INTL 
\$52.58 } \\

Action:Think[{"response": "The third product matches the specific requirements. I should click the third product."}] \\

Observation: OK \\

Action:click[{"button": "B073KR5D4W"}] \\

Observation: WEB PAGE: {

[Back to Search] 

[Prev] 

size [1ft][2ft][3ft][4ft][5ft][6ft][8ft][10ft][12ft][15ft][20ft][25ft][25 ft][30ft][35ft][40ft][45ft][50ft][55ft][60ft][65ft][70ft][75ft][80ft]

[85ft][90ft][95ft][100ft][105ft][110ft][115ft][120ft][125ft][130ft][135ft][140ft][145ft][150ft][155ft][160ft][165ft][170ft][175ft][180ft]

[185ft][190ft][195ft][200ft][210ft][220ft][230ft][240ft][250ft][260ft][270ft][280ft][290ft][300ft][310ft][320ft][330ft][340ft][350ft]

[360ft][370ft][380ft][390ft][400ft]

style [quadshield - black][quadshield - white][quadshield solid copper - black][solid copper core - black][solid copper core - white][solid copper w |  weather boot - black][solid copper w |  weather boot - white][trishield - black][trishield - white][usa made trishield - black][usa made trishield - white][direct burial - black][direct burial - orange][direct burial 3ghz - orange][direct burial 3ghz w |  weather boot - ora...][direct burial rg-11 connector - orange][direct burial rg-11 w |  weather boot - or...][direct burial w |  rg-11 connector - black][direct burial w |  rg-11 weather boot - bl...][direct burial w |  weather boot - black][direct burial w |  weather boot - orange][plenum rg-11 - white][w |  ground - black][w |  ground - white][3ghz dual w |  ground - black][3ghz w |  messenger - black][dual - black][dual copper w |  ground - black][dual copper w |  ground - white][w |  messenger - black]
135ft Black Quad Shield Solid Copper 3GHZ RG-6 Coaxial Cable 75 Ohm (DIRECTV Satellite TV or Broadband Internet) Anti Corrosion Brass Connector RG6 Fittings Assembled in USA by PHAT SATELLITE INTL 
Price: \$52.58 
 
[Buy Now] } \\

Action:click[{"button": "20ft"}] \\

Observation: WEB PAGE: {You have clicked 20ft.} \\

Action:click[{"button": "quadshield solid copper - black"}] \\

Observation: WEB PAGE: {You have clicked quadshield solid copper - black.} \\

Action:click[{"button": "Buy Now"}] \\

Observation: shopping is finished.

\end{tcolorbox}
\newpage

\end{document}